\documentclass{article}

\usepackage{arxiv}

\usepackage[utf8]{inputenc} 
\usepackage[T1]{fontenc}    
\usepackage{hyperref}       
\usepackage{url}            
\usepackage{booktabs}       
\usepackage{amsfonts}       
\usepackage{amsmath}
\usepackage{nicefrac}       
\usepackage{microtype}      
\usepackage{graphicx}
\usepackage[numbers]{natbib} 
\usepackage{doi}
\usepackage{pifont}
\usepackage[ruled,vlined]{algorithm2e}
\usepackage{multirow}

\title{RoBus: A Multimodal Dataset for Controllable Road Networks and Building Layouts Generation}

\author{ {\hspace{1mm}Tao Li, Ruihang Li, Huangnan Zheng, Shanding Ye, Shijian Li, Zhijie Pan\thanks{Corresponding Author. The authors' emails are \{litaocs, 12221089, hnz, ysd, shijianli, zhijie\_pan\}@zju.edu.cn.}} \\
	Department of Computer Science and Technology\\
	Zhejiang University\\
	\\
}

\renewcommand{\shorttitle}{RoBus: A Multimodal Dataset for Controllable Road Networks and Building Layouts Generation}

\begin{document}
\maketitle

\begin{abstract}
Automated 3D city generation, focusing on road networks and building layouts, is in high demand for applications in urban design, multimedia games and autonomous driving simulations.
The surge in generative AI models facilitates the design of city layouts in recent years. 
However, the lack of high-quality datasets and benchmarks hinders the progress of these data-driven methods in generating road networks and building layouts.
Furthermore, few studies consider urban characteristics, which are generally analyzed using graphics and are crucial for practical applications, to control the generative process.
To alleviate these problems, we introduce a multimodal dataset with accompanying evaluation metrics for controllable generation of \textbf{Ro}ad networks and \textbf{Bu}ilding layout\textbf{s} (\textbf{RoBus}), which is the first and largest open-source dataset in city generation so far. 
RoBus dataset is formatted as images, graphics and texts, with $72,400$ paired samples that cover around $80,000 \, km^2$ globally.
We analyze the RoBus dataset statistically and validate the effectiveness against existing road networks and building layouts generation methods. 
Additionally, we design new baselines that incorporate urban characteristics, such as road orientation and building density, in the process of generating road networks and building layouts using the RoBus dataset, enhancing the practicality of automated urban design.
The RoBus dataset and related codes are published at \url{https://github.com/tourlics/RoBus_Dataset}.
\end{abstract}


\begin{figure}
  \includegraphics[width=\textwidth]{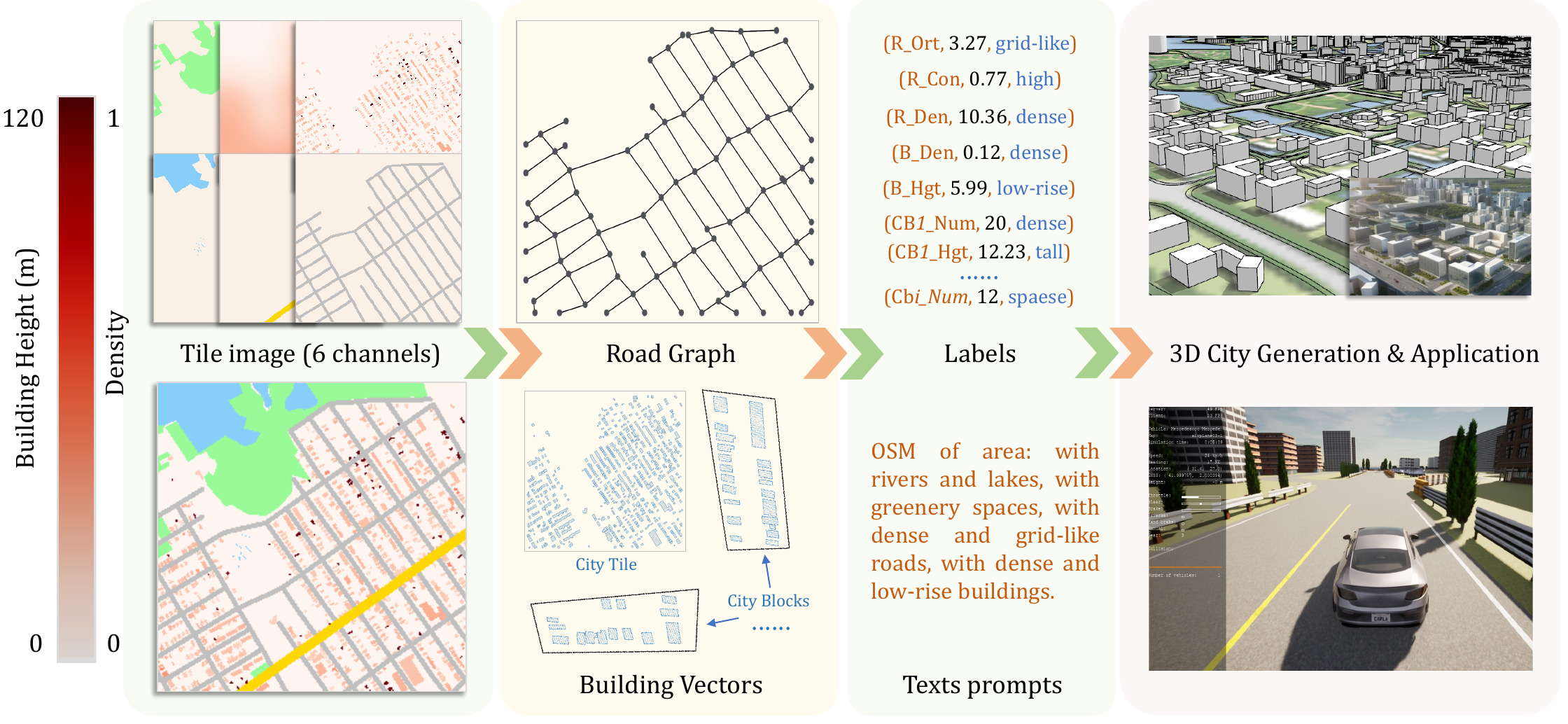}
  \caption{An example from the RoBus Dataset, including images, graphics, and texts that describe road networks and building layouts in a city tile. The RoBus Dataset is scalable for various 3D urban generation tasks and practical applications.} 
  \label{fig:teaser}
\end{figure}

\section{Introduction}
The generative design of three-dimensional (3D) cities is of considerable importance across multiple fields such as urban planning \cite{zheng2023road, liao2024generativeai}, multimedia games \cite{vimpari2023}, and autonomous driving simulations \cite{tian2024occ3d}. 
Road networks and building layouts are core components in designing or generating 3D urban configurations.
The manual design of these components has been prevalent in both the game industry and urban planning for a long time, which is often criticized for being costly expensive and time-consuming.
Consequently, there is a growing demand for automated approaches to generate diverse and extensive urban layouts tailored to specific city characteristics, which has prompted a significant increase in research on automated urban generation recently. 
Such research not only alleviate the limitations of manual design but also deepens the understanding of our living environments.

Traditional procedural modeling methods progressively generate road networks or building layouts based on grammars or L-systems \cite{parish2001procedural, lechner2003procedural, lechner2006procedural}, which rely on expert knowledge to manually design certain rule sets.
With the rapid advancement of generative artificial intelligence (AI), new approaches employing deep generative models have emerged in numerous design tasks. 
Highly relevant and prevalent research topics include the generation of poster layouts and house floor plans \cite{inoue2023layoutdm, xu2021blockplanner}, where numerous data-driven methods have been proposed to produce varied and contextually appropriate results, significantly reducing the reliance on extensive domain knowledge \cite{wang2022controllable}.
Overall, research on these small-scale (concerning the size of entities) layout generation has become mature due to the availability of numerous public datasets \cite{zheng2019magazine, wu2019rplan}.
In contrast, methods for large-scale urban layout generation remain relatively immature. 
Existing works often rely on a limited amount of self-collected data (generally private) for specific tasks, such as road completion \cite{fang2022incorporating} and buildings generation \cite{wu2022ganmapper}.
Obviously, the lack of adequate, high-quality, open-source datasets and benchmarks significantly hinders the progress of data-driven methods for city-scale road networks and building layouts.
Furthermore, few existing studies consider urban characteristics, where graphics such as road network typologies and vectored buildings are generally expected, during the generative process. 
This oversight of graphic data and urban features results in existing methods being sub-optimal for synthesizing new road networks and building layouts with desired properties. These capabilities are crucial for practical applications in fields like urban planning and the game industry.

To address these challenges and support the advancement of techniques for automated 3D city generation, we introduce the \textbf{RoBus} dataset for controllable \textbf{Ro}ad Networks and \textbf{Bu}ilding Layout\textbf{s} Generation. 
As shown in Fig. \ref{fig:teaser}, the RoBus dataset includes images, graphics and texts, providing a comprehensive description of 3D urban layouts, which is the first and largest multimodal dataset in generative 3D urban designing.
The RoBus dataset contains $72,400$ paired samples that cover approximately $80,000 km^2$ of different places across the world, showcasing remarkable diversity.
Additionaly, the RoBus dataset is scalable for various tasks related to 3D urban generation, such as geometry constrained city layouts generation or competition, road graph generation, vectored building layouts generation, text to image generation and so on.
To validate the effectiveness of the RoBus dataset, we apply prevalent methods for road network and building layout generation using the dataset. Besides, we establish a comprehensive benchmark to assess the quality, diversity, validity and urban properties of generated results.
More importantly, we propose a baseline that integrates road attributes into generative models to synthesize desired road layouts, and enhances the traffic convenience by concentrating on topological structure of generated road networks.
For the building layouts generation, we design a baseline that directly generate vectored buildings with height information, by incorporating building attributes (shown in Fig. \ref{fig:attr}) into latent space in the model to generate building layouts conditioned on the user-defines density.
To validate the applicability of the RoBus dataset and proposed baselines, we apply the generated road networks and building layouts in \textit{Carla}\cite{dosovitskiy2017carla}, which is designed for autonomous driving simulations based on the widely used \textit{UnrealEngine} in game industry.

The proposed RoBus Dataset is characterized by its diversity, scalability, usability, and applicability. Our contributions can be summarized as follows:

\begin{itemize}
    \item We release RoBus dataset, the first and largest multimodal dataset in generative 3D urban design, including images, texts and graphics such as topological road networks and vectored building with height.
    
    \item We propose two baselines that incorporate urban characteristics, such as road orientation and building density, into the generative process of deep-learning models based on the RoBus dataset.
    
    \item We establish a benchmark to evaluate existing generative methods for road networks and building layouts in aspect of quality, diversity, validity and urban properties of synthesized results.
    
    \item Experiments on prevalent generative tasks related to 3D urban generation demonstrate the usability and scalability of the RoBus dataset. We also extend the generated results into game engines to showcase the application values.
\end{itemize}

\section{Related Works}
Computer-aided urban design has been a significant focus of computer graphics research for many years. This field has witnessed a transition from traditional procedural modelling to data-driven generative AI models.
Traditional methods \cite{parish2001procedural, lechner2003procedural} depend on expert knowledge to manually create rules for generating roads, buildings or terrain.
With the recent increase in deep generative models, more researchers are exploring the potential of these models for urban design. 
In this section, we briefly review these AI-based methods for designing road networks and building layouts. Then we summarize the datasets relevant to these data-driven approaches.

\subsection{Road Networks Generation Methods}
Methods based on generative AI models usually learn the distribution of road networks from real world, free from the dependency on extensive domain knowledge. 
These methods can be generally divided into image-based and graph-based approaches.
\textbf{Image-based road network generation} approaches treat road network generation as an image generation problem. These methods utilize models such as generative adversarial networks (GANs) or variational autoencoders (VAEs) to learn the pixel-level distribution of road networks within images.
Hartmann et al. first proposed the GAN-based road network generation pipeline StreetGAN \cite{hartmann2017streetgan}.
Subsequently, numerous studies \cite{kelvin2020cgan, fang2022incorporating, yang2023street} have attempted to use conditional GANs to generate roads according to some user-defined inputs or contextual content. 
Besides GAN-based models, Murcio et al. \cite{kempinska2019modelling} trained a VAE model to capture real-world road patterns and generate new roads by sampling the encoded features.
To solve the problem of generating large-scale road networks, Birsak et al. \cite{birsak2022large} introduced the Variational Quantized VAEs into large-scale urban road network generation by incorporating population constraints.  
More recently, Przymus et al. \cite{przymus2023map} and Qin et al. \cite{qin2024text2city} applied diffusion models to generate maps via text prompts.
\textbf{Graph-based road network generation} approches model the road network as a topological planar graph consisting of nodes and edges, framing the road network generation as a planar graph generation problem.
Inspired by traditional turtle graphics methods, Chu et al. \cite{chu2019neural} proposed the Neural Turtle Graphics (NTG) model for generating road graphs, which employs an encoder-decoder with a recursive neural network to process and generate road networks. 
However, it is primarily limited to understanding patterns within a specific local scope and is suited only for small-scale road network generation.
To address this limitation, Owaki et al. \cite{owaki2020roadnetgan} introduced RoadNetGAN, applying the principles of NetGAN \cite{bojchevski2018netgan} to large-scale road network generation using random walks. However, it struggles to produce diverse and plausible road configurations, often resulting in numerous sharp turns.
Overall, Image-based approaches for road networks generation suffer from modeling topological features, while graph-based approaches struggle to encode spatial information.
Furthermore, both of them fail to generate road networks with the desired properties.

\subsection{Building Layouts Generation Methods}
Research on building layouts generation have certain similarities with generating road network layouts, as both learn spatial patterns from real-world design instances.
Models like LayoutGAN++ \cite{kikuchi2021constrained} and LayoutVAE \cite{jyothi2019layoutvae} provide basic frameworks for layout generation, which are widely used and improved in tasks such as document and poster layouts generation \cite{zhang2023layoutdiffusion, qu2023layoutllm, chai2023two}, house floor planning \cite{chang2021building, nauata2020house}.
When it comes to larger-scale building layouts generation, 
Fedorova et al. \cite{fedorova2021generative} and Quan et al. \cite{quan2022urban} use GAN-based methods to generate building layouts for various cities, demonstrating the model's adaptability to different urban morphologies.
BlockPlanner \cite{xu2021blockplanner} is proposed for generating building layouts in dense urban blocks, assuming buildings are in rectangular shapes within rectangular city plots, and uses graph models to manage the building layouts. 
BlockPlanner is limited for more complex block shapes, irregular buildings, and large-scale building layouts.
To address these limitations, He et al. \cite{he2023globalmapper} introduced a VAE based graph attention network for modelling and generating building layouts, capable of handling arbitrary block shapes and building types.
Additionally, generative AI design for building layouts has widely attracted experts on urban designing \cite{liao2024generativeai, jiang2023building, wang2023automated}, which prompts the application in urban analysis and planning.

\subsection{Datasets for Generative Urban Design}
Generally, small-scale layout generation tasks such as poster or indoor room layout are relatively mature and widely studied, benefiting from numerous public dateset such as Magazine \cite{zheng2019magazine} and RPLAN \cite{wu2019rplan} dataset, which collects 3,919 magazine pages and 80K room layouts respectively.
For larger-scale generation tasks such as urban road networks and building layouts generation, it have been a long time that struggling with limited amount of data. For instance, Yan et al. \cite{yan2019graph} collect $2,194$ samples to train a model that classify building patterns. Wu et al. \cite{wu2022ganmapper} collect data from Singapore to generate building layouts based on road networks. Przymus et al. \cite{przymus2023map} provide large amount raw maps for text to image generation. Thanks to recent works published by Chen et al. \cite{chen2023reco} and He et al. \cite{he2023globalmapper}, which collects large amount of data for community building layouts generation, prompts the advancement of large-scale generation methods.
However, these open-source data lacks of the city-scale buildings and matched road networks, which has a gap to applied to large-scale city generation.
Situations in road network generation are even worse.
Researchers on this area are struggling with inadequate data samples. 
For example, Chu et al. \cite{chu2019neural} collect  $17$ unique cities and select the most densely annotated
$10km^2$ region within each city from OSM by complex preprocessing. Owaki\cite{owaki2022road} et al. collect 4 cities from Japan of $10km^2$.
Yang et al. \cite{yang2023street} collect $2586$ road images from three major Australian cities. Birsak et al.  \cite{birsak2022large} collect around $400 km^2$ datas.
Unfortunately, these self-collected data are of small amount and are close-sourced. Research in road networks generation are eaging for a large-scale dataset and benchmarks.

\textbf{To summary}, numerous datasets for poster and house floor plans are benefiting these small-scale layout generation tasks. 
Recent studies that focus on larger scale generation tasks, such as community layout planning, often provide datasets that only include buildings. These datasets are insufficient for city generation, as they lack data on city-scale buildings and corresponding matched road networks.
Moreover, research on road network generation typically relies on self-collected data comprising a limited number of samples. This highlights a pressing need for large-scale, high-quality road network datasets and corresponding benchmarks.
In this paper, we intend to overcome existing challenges in 3D urban generation through the release of the RoBust Dataset, which is expected to encourage progress in data-driven methods for generating road networks and building layouts.

\begin{figure}[th]
  \centering
  \includegraphics[width=\linewidth]{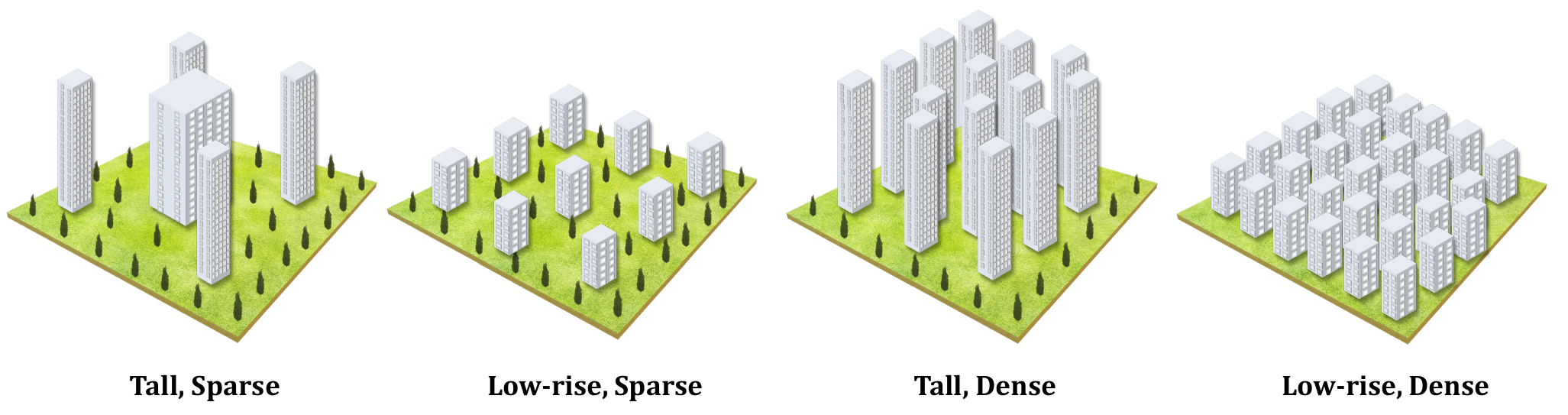}
  \caption{Considering urban characteristics, we classify buildings in a city block according to density and average height.}
  \label{fig:attr}
\end{figure}

\begin{figure*}[ht]
  \centering
  \includegraphics[width=\linewidth]{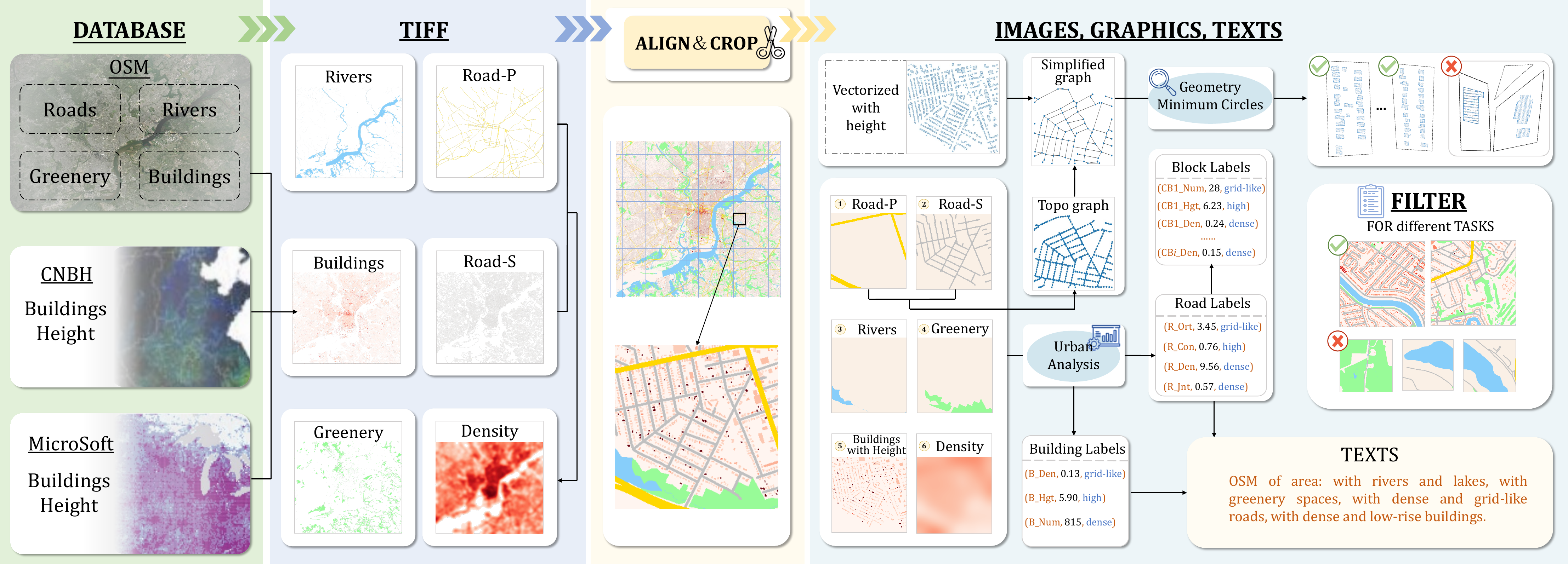}
  \caption{The Generating Pipeline of RoBus Dataset.} 
  \label{fig:pipeline}
\end{figure*}

\section{RoBus Dataset}
In this section, we firstly describe the proposed RoBus dataset, and then illustrate the pipeline to generate the dataset from raw data. Lastly, we analyze and explore the RoBus dataset statically and demonstrates its highlights.  

\subsection{Dataset Description}
The RoBus dataset is a comprehensive and multimodal dataset for generative urban design, focusing on generation of road networks and building layouts. It is composed of:

\textbf{\textit{Images}}: This component includes \textit{tif} files with 6 channels, each representing a different urban element: primary roads, secondary roads, water bodies, green spaces, buildings with heights, and density. Each image contains coordinate reference information, which facilitates further processing and expansion.

\textbf{\textit{Graphics}}: This component includes simplified road network graphs in \textit{gpickle} format and vectored buildings with heights in \textit{geojson} format. These 3D vectored buildings are provided at both city tile and block scales.

\textbf{\textit{Texts}}: This component encompasses statistical values, labels, and descriptive texts that detail characteristics of urban tiles, such as road orientation, traffic convenience, building height, and density.

The RoBus dataset encompasses an extensive area, covering approximately $80,000$ square kilometers across multiple regions in Australia, China, Europe, and the United States. This broad geographic coverage ensures that the dataset captures a diverse range of urban layouts.
It is designed to be scalable for a variety of existing generative tasks related to road networks and building layouts, including:
\textbf{Geometry Constrained Generation} such as generating secondary roads based on primary road networks, roads based on density and land-use maps \cite{yang2023street}, building layouts based on road layouts \cite{wu2022ganmapper}, or boundary maps \cite{chen2023reco}.
\textbf{Graphic Generation} such as generating topological road networks \cite{chu2019neural, owaki2022road} and vectored building layouts \cite{he2023globalmapper}.
\textbf{Text-to-Image Generation} techniques, such as urban map generation \cite{przymus2023map} and completion \cite{qin2024text2city}, utilize text descriptions with popular models like CLIP and Stable Diffusion.
\textbf{Others} including urban analysis and planning \cite{boeing2019urban}, road network competition \cite{fang2022incorporating}, and road topology extraction \cite{li2023topology}.

\subsection{Data Collection and Generating Pipeline}
As shown in Fig. \ref{fig:pipeline}, our data collection and generating pipeline can be generally summarized as follows:

\textbf{Stage 1: Data Collection and Preprocessing.}
To capture the spatial patterns of road networks and building layouts on a large scale, we collect raw data from OpenStreetMap \footnote{\url{https://extract.bbbike.org/} \label{osm}} to extract road networks, rivers, greenery and building contours by filtering different OSM tags, which are detailed in the appendix. 
However, data from OSM are generally noisy and incomplete, and thus cannot be directly applied to model training without complex preprocessing \cite{chen2023reco}. To clean the road network data, we rasterize roads into images and then apply thinning to modify the roads, as suggested by \cite{wu2022ganmapper}.
Additionally, we observed that building contours from OSM frequently lack height information and are often incomplete. To address these limitations, we enhanced the extracted OSM data with publicly available datasets regarding building heights. Notably, we collected data from Microsoft\footnote{\url{https://planetarycomputer.microsoft.com/dataset/ms-buildings} \label{ms}} for areas in Australia, Europe, and the United States, as well as the CNBH dataset \cite{wu2023cnbh} for areas in China.
We align OSM building contours with CNBH tiles and calculate the average height values of valid pixels. For areas with missing values, we estimate the heights using nearby buildings within a $300m$ radius. If data are still unavailable, we default to a height of $24$ meters.
Through this meticulous preprocessing, we aimed to enhance the quality and reliability of the building information in the RoBus dataset, which is crucial for accurate urban modeling and generation.

\textbf{Step 2: Construction of Images.}
To construct the image component of the RoBus dataset, we rasterize the preprocessed data to produce \textit{tif} files with six channels. These channels include primary roads (Road-P), secondary roads (Road-S), water bodies (rivers or lakes), green spaces, density \cite{birsak2022large}, and buildings with height information.
We choose the Pseudo-Mercator projection (EPSG:3857) as coordinate reference system (CRS) to these tif images, with a resolution of 5 meters per pixel.
Given that each \textit{geojson} file may have slightly different boundaries when rasterized, we identified the common area and align them to ensure consistency.
After alignment, we crop the \textit{tif} images into small tiles with a 20\% overlap, each sized at $256 \times 256$ pixels, corresponding to an actual area of approximately $1.6 km^2$.
These tiles with $6$ channels constitute the image part of the RoBus dataset, which supports to various image-based generation tasks.
Additionally, we preserve the coordinate system and geographic coordinates during the cropping process, aiming to facilitate further research on larger-scale generation tasks.

\textbf{Step 3: Construction of Graphics.}
To construct the topological graph of road networks, we combine the first two channels of the images, which represent primary and secondary road networks, and then skeletonize the result using morphological thinning methods.
Given that skeletonized road graphs are densely populated with redundant nodes, leading to excessive computational costs for further processing, we simplify these graphs using Eq. \ref{eqn: eqgraph2}. This simplification process preserves critical features such as road joints and sharp turns. In Eq. \ref{eqn: eqgraph2}, $\mathbf{e}_{ki}$ represents the vector from node $v_k$ to node $v_i$ in the road graph $G_R$. The function $m(\cdot)$ refers to the operation that merges two closely positioned nodes, and $C{tr}$ is the threshold defined for smoothness.
The simplified road network graph serves three primary purposes. First, it facilitates the computation of topological attributes essential for transportation research, crucial for generating the text domain in our dataset. Second, the topology of road networks typically forms cycles that outline city blocks. Third, the graphic structure is directly applicable to topological road generation methods like random walks \cite{chu2019neural}.

\begin{equation}
\label{eqn: eqgraph2}
    C_{tr} < |\cos{(\mathbf{e}_{ki} \cdot \mathbf{e}_{kj})}| \leq 1, v_k \in \{v \in m(\mathbf{G_R}) | deg(v)=2 \} 
\end{equation}

We provide vectored building layouts with heights at both the city tile scale and the city block scale.
To generate building layouts at the block scale, we use boundaries that are automatically partitioned by road networks.
A critical challenge in this process is to efficiently identify as many geometric minimal cycles within the road network's topology as possible.
For clarity, we define the \textit{geometric minimal cycle} as a cycle in a graph with geometric coordinates that does not enclose any other cycles. Most existing algorithms primarily focus on finding basic cycles or enumerating all cycles within a topological graph, without considering the geometric positions.
To address this gap, we develop a new algorithm tailored for finding geometric minimal cycles. 
Initially, we iteratively remove all vertices with a degree of one. Next, for each node with a degree of two, we identify simple paths that include the node and its two immediate neighbors, restricting the path length to a maximum of 12 edges to ensure efficiency. Finally, we examine all simple paths to pinpoint the shortest cycles, designated as geometric minimal cycles. We repeat the last two steps until no vertices remain in the graph. Details can be found in the appendix.

The topological graph of road networks and the vectored buildings with heights at both the city tile scale and the city block scale constitute the graphics component of the RoBus dataset. These graphics are crucial for modeling and generating urban layouts with specific properties, essential for advanced 3D urban analysis and planning.

\textbf{Step 4: Construction of Texts.}
We analyze the attributes of the road networks and building layouts statically to generate labels and texts.
Drawing on urban planning research, we categorize the road networks based on their density and orientation.
Road density is classified as either 'dense' or 'sparse', influencing traffic flow and accessibility. The orientation, assessed through the entropy of street bearings \cite{boeing2019urban}, indicates the degree of order (e.g., grid-like road networks) or disorder (e.g., random road networks) in the road layout.
We also categorize buildings based on density and average height, which includes categories illustrated in Fig. \ref{fig:attr}.
On basis of these labels and categories, we generate descriptive text using predefined templates, where sentences begin with "OSM," as suggested by \cite{przymus2023map}. These texts serve as prompts to describe the characteristics of urban areas and hold potential for application in text-to-image models.

\textbf{Step 5: Filter for Different Tasks.}
For specific generation tasks like creating building layouts from road networks, we filter out tiles that lack roads or buildings. This filtering process is similarly applied to other targeted tasks.

\subsection{Dataset Analysis and Highlights}

\begin{table}[!tb]
\centering
\caption{Statistics of RoBus Dataset.}
\begin{tabular}{cccc}
  \toprule
  Stats & Road Length ($km$)      & \# of Buildings      & \# of Boundries         \\ 
  \midrule
  Max   & 35.13 & 4196 & 66 \\
  Mean  & 8.64        & 312.10       & 6.38         \\
  Total & 625,944          & 22,596,169       & 461,666        \\ 
  \bottomrule
\end{tabular}
\label{tab:stats}
\end{table}

\begin{figure}[tb]
  \centering
  \includegraphics[width=\linewidth]{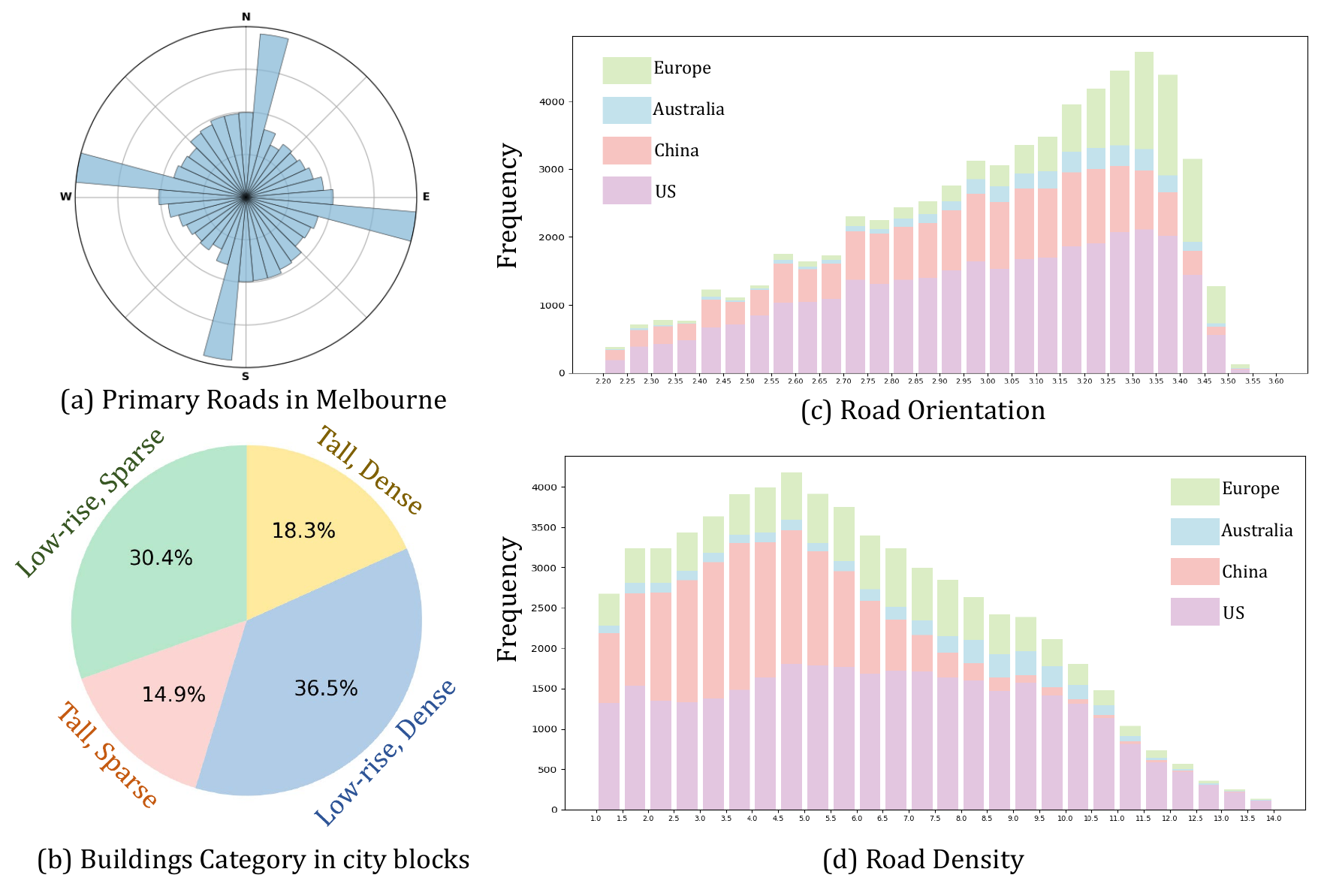}
  \caption{Visualization of Statics of RoBus Dataset. }
  \label{fig:statics}
\end{figure}

To provide a more detailed demonstration of the RoBus dataset, we perform a static analysis of key urban elements such as road length, building count, and boundary count at the city tile scale, detailed in Table \ref{tab:stats}.
To enhance the visualization of these static results, we present them in Fig. \ref{fig:statics}. Specifically: (a) illustrates the orientation of primary roads in Melbourne as selected from the RoBus dataset.
(b) displays the proportion of different building types within a city block.
(c) and (d) show the distribution of road orientation and road density in the RoBus dataset, respectively.
We compare the RoBust dataset with existing relevant dataset, as shown in Tab. \ref{tab:comparision}. 
The RoBus dataset is the largest of its kind with $72,400$ paired samples that include multimodal data such as road graphs, building vectors, and labels, which are missing in existing datasets.

\begin{table*}
\scriptsize
\centering
\caption{Comparision with exisiting datasets}
\label{tab:comparision}
\begin{tabular}{ccccccc} 
\toprule
Dataset/Papers    & Accessibility & Covering(Samples\#)            & Road Graph     & Building Vec    & Labels & Tasks                        \\ 
\midrule
NTG \cite{chu2019neural}              & Closed        & $170km^2   $                 &   \ding{52}            & \ding{56}               & \ding{56}         & Road Graph Generation        \\
Owaki et al. \cite{owaki2020roadnetgan}            & Closed        &       $11.56 km^2$      & \ding{52}               & \ding{56}       & \ding{56}             & Road Graph Generation        \\
Fang et al. \cite{fang2022topography}         & Closed        &    $\#56,000$          & Image          & \ding{56}         & \ding{56}            & Road Network Completion  \\
Birsak et al. \cite{birsak2022large}             & Closed        &       $400 km^2$   & Image          & \ding{56}           & \ding{56}            & Road Image Generation        \\
GanMapper \cite{wu2022ganmapper}         & Closed        &    $\#12,139$          & Image          & \ding{56}         & \ding{56}            & Building Layouts Generation  \\
Reco \cite{chen2023reco}             & \textbf{OPEN}          & $\#37,646$ & \ding{56}          & \ding{52}                & \ding{56}              & Community Layouts generation  \\
\textbf{RoBus}             & \textbf{OPEN}         & $80000km^2(\#72,400)$          & \ding{52}  & \ding{52}  & \ding{52}                 & 3D Urban Generation          \\
\bottomrule
\end{tabular}
\end{table*}

In summary, our dataset is highlighted as: 
\textbf{Multimodal}: The dataset delivers a comprehensive multimodal description of urban scenes, integrating images, graphics, and texts. The strong correlation among these modalities guarantees a synchronized perspective of urban layouts.
\textbf{Diversity}: The dataset covers approximately $80,000 km^2$ across various locations worldwide, showcasing extensive diversity.
\textbf{Scalability}: The dataset supports a variety of tasks relevant to 3D urban generation, including geometry constrained image generation, graphics generation and text to image generation. 
\textbf{Usability}: Experiments have been conducted on multiple tasks using this dataset, demonstrating its effectiveness and usability. 
\textbf{Applicaplity}: The dataset accompanying with proposed generation baseline can be readily applied to 3D games.
We expect the RoBust dataset to encourage progress of data-driven methods for 3D urban design.

\section{Experiments}
\label{sec:exper}
In this section, we design and conduct experiments to address the following research questions (RQs) both qualitatively and quantitative using the proposed benchmarks detailed in Section \ref{sec:results}.
We target at demonstrating the effectiveness, scalability, and applicability of the RoBus dataset, as well as the proposed baselines that incorporate urban attributes into the generative design process.

\textit{RQ1: Can the dataset be applied to existing road networks and building layouts generation methods conditioned on geometric constraints, and how does it perform?}

\textit{RQ2: Can the dataset be applied to solve the proposed tasks that generating urban layouts with desired properties? Does it make any difference to the generated results?}

\textit{RQ3: Are the generated results applicable for use in 3D games or autonomous driving simulations?}

\textit{RQ4: How does the image resolution, size and distribution of the dataset affect the quality of the generated results?}

\subsection{Generation based on Geometry Constraints}
We select two representative tasks, generating road networks conditioned on landuse maps (\textbf{Task \uppercase\expandafter{\romannumeral1}}) and generating building layouts conditioned on road networks (\textbf{Task \uppercase\expandafter{\romannumeral2}}), to conduct the geometry constraints based urban generation.
GANs have marked a significant advancement in generating road networks \cite{hartmann2017streetgan,yang2023street} and building layouts \cite{wu2022ganmapper, wu2023instantcity, jiang2023building}. 
Given the widespread adoption of GANs, we trained the popular model pix2pix \cite{isola2017image} for Task \uppercase\expandafter{\romannumeral1} and \uppercase\expandafter{\romannumeral2}.

\subsection{Generating with Desired Properties}
\label{sec:attribute}

\begin{figure}[h]
  \centering
  \includegraphics[width=0.8\linewidth]{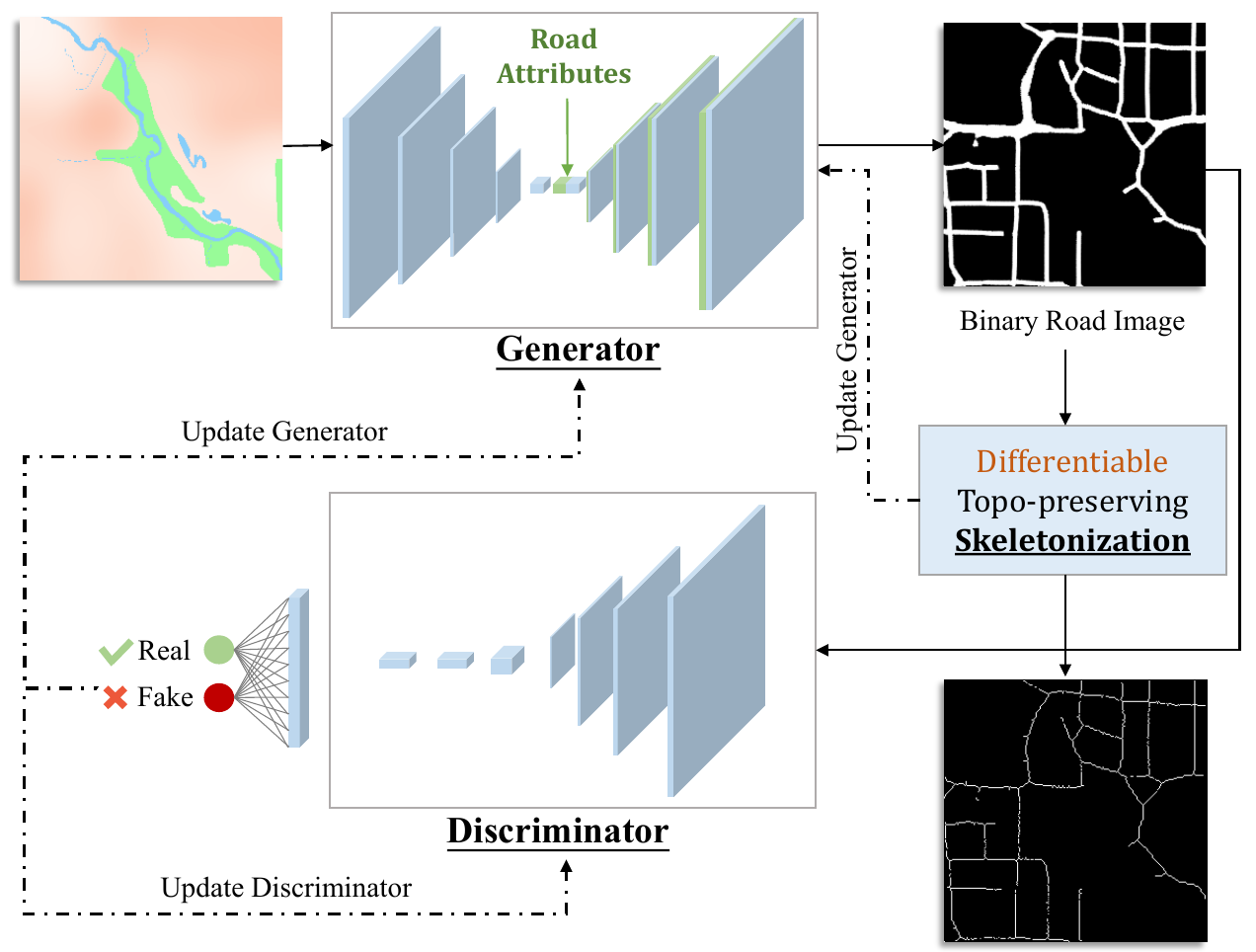}
  \caption{Framework of Task \uppercase\expandafter{\romannumeral3}.}
  \label{fig:toporoads}
\end{figure}

For the task of generating road networks (\textbf{Task \uppercase\expandafter{\romannumeral3}}), we introduce a baseline model that incorporates desired characteristics, such as grid-like road networks with low orientation order, and optimize road connectivity to achieve higher traffic convenience.
To construct the simplest version of our baseline, we adapt the pix2pix model to constitute the backbone, as shown in Fig \ref{fig:toporoads}.
The channel number of input local density maps and the targeting road network maps in the generator are limited to one.
We integrate the road attributes $R$ such as global road density and orientation order in the encoded latent vector in the U-like generator.
The decoder synthesize binary road images conditioned on the encoded local density maps and the concatenated road attributes vectors.

To enhance the topology of the generate roads, which have an great impact on the urban attribute of traffic convenience, we focus on the topological structure of generated roads.
Specifically, we extract the topological skeleton of the synthesized images and calculate the center-line dice score to enhance the road connectivity, which has proven to be differentiable in \cite{shit2021cldice, menten2023skeletonization}. 
The overall loss of the generator is formulated as Eq. \ref{eqn: topopix}.

\begin{equation}
\label{eqn: topopix}
    \mathcal{L}(G_\theta) = \lambda_1 \mathcal{L}_1 + \lambda_2 \underset{p_{x}(z|r)}{\mathbb{E}}[-\log D(G_\theta (z|r)] + \lambda_3 \mathcal{L}_{topo}
\end{equation}

where $\mathcal{L}_{1}$ is the L1 loss to evaluate the similarity of the generated image and the ground-truth images, $r$ is the road attributes conditions, $\mathcal{L}_{topo}$ is the center-line dice loss to enhance the topology of generated graph roads.

\begin{figure}[h]
  \centering
  \includegraphics[width=\linewidth]{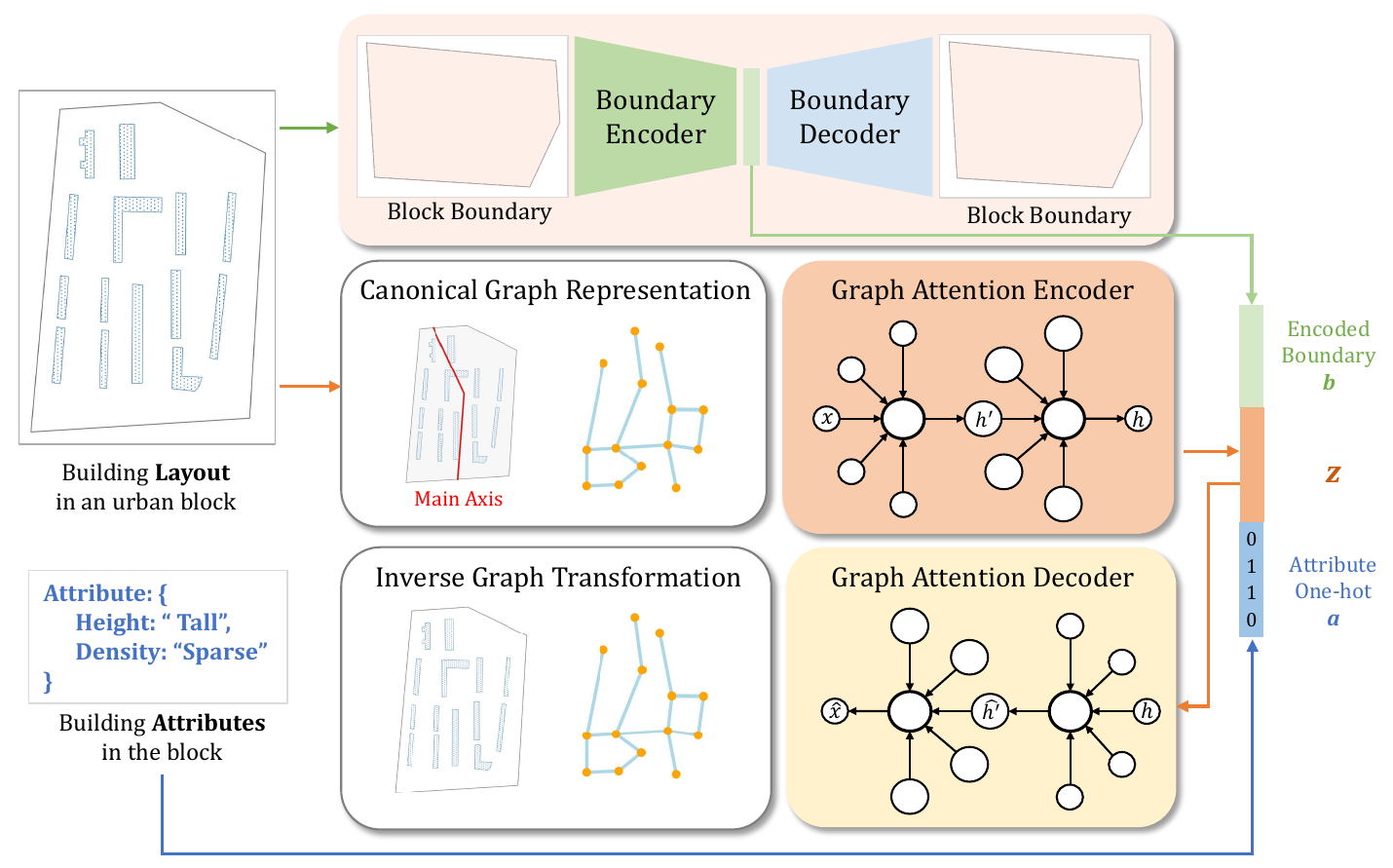}
  \caption{{Framework of Task \uppercase\expandafter{\romannumeral4}.}}
  \label{fig:vecbuilding}
\end{figure}

For vectored building layout generation tasks, we focus on generating buildings layouts constrained by the boundary of a city block (\textbf{Task \uppercase\expandafter{\romannumeral4}}), which match the predetermined density and heights.
This is achieved by integrating building attributes into the generative models Conditional Variational Autoencoders (cVAEs).
As shown in Fig. \ref{fig:vecbuilding}, we includes an autoencoder to compact the building boundaries into latent vectors $b$, and graph attention networks (GAT) \cite{velickovic2018graphatten} as the backbone of the encoder and decoder.
Additionally, we encode the building heights and density as one-hot vector $a$ to serve as the attribute prior of buildings in the city block.
We follow GlobalMapper \cite{he2023globalmapper} to transform buildings into a canonical spatial format and subsequently into graph structures $G$.
To enable the learning of building heights, we add building heights to nodes' attributes in $G$, as well as the original building location and minimum bounding box in GlobalMapper \cite{he2023globalmapper}.

During the training process, the conditional VAE learns to capture the distribution $p(G_B|b,a)$ in the dataset, which represents the probability of generating a building graph $G_B$ conditioned on the encoded boundary vectors $b$ and attribute vectors $a$. The learned distribution is then sampled for generating new building graphs.
The model captures $p(G_B|b,a)$ by maximizing its evidence lower bound, as frequently used in conditional VAEs \cite{zhu2019elbomm, he2023globalmapper}.
Additionally, to make the model focusing on building attributes such as heights, we measure the similarity with groundtruth $\hat{G_B}$ using L2 loss.
To sum up, the overall loss functions is formulated as Eq. \ref{eqn: vae}.

\begin{equation}
\label{eqn: vae}
    \mathcal{L} = \beta_1 ||G_B^h - \hat{G_B}^h||_2 + \beta_2 [\underset{q}{\mathbb{E}}(log(p(G_B|z,b,a)))-D_{KL}(q|| p(z|b,a)) ]
\end{equation}

where $h$ refers to the attributes in the graph such as heights, $p(z|b,a))$ is the prior distribution of $z$ conditioned $b$ and $a$, and $q$ refers to $q(z|G_B,b,a)$, which is the approximate posterior distribution of the latent variables.

\section{Results and Analysis}
\label{sec:results}
In this section, we introduce the benchmarks with comprehensive evaluation metrics. 
Additionally, we report our qualitative, quantitative and ablation results to answer the RQs in section \ref{sec:exper}.

\subsection{Evaluation Metrics}
To comprehensively evaluate methods related to the generation of road networks and building layouts, we introduce the benchmark that assesses the quality, diversity, validity, and urban properties of the synthesized results.

\begin{figure}[h]
  \centering
  \includegraphics[width=\linewidth]{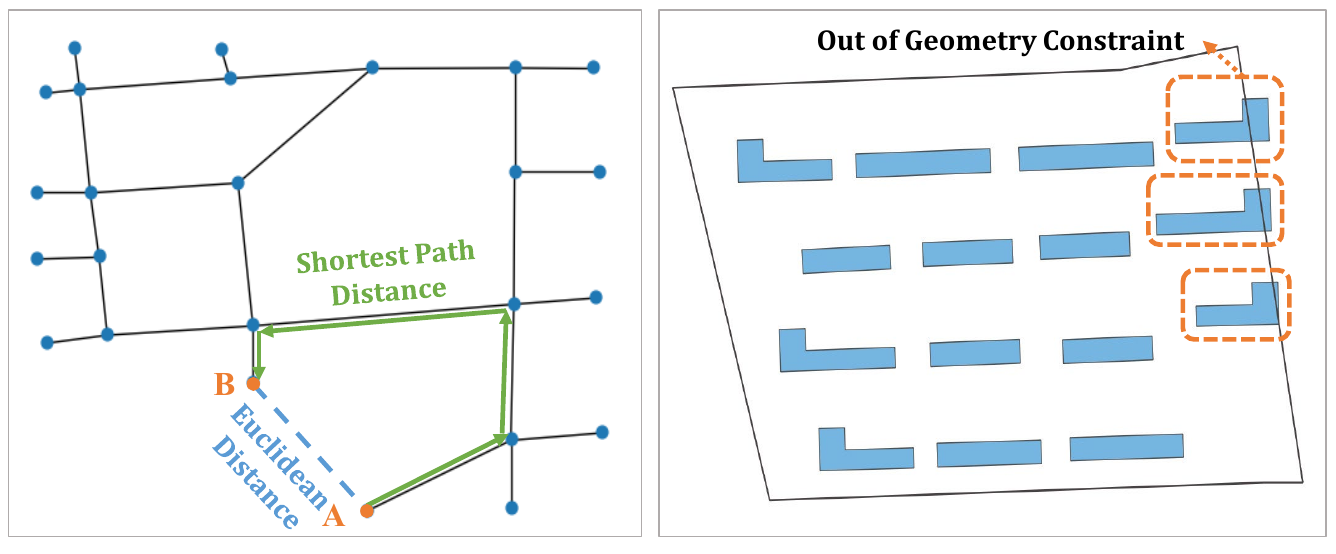}
  \caption{Illustration of Traffic Convenience and Validity.}
  \label{fig:metrics}
\end{figure}

\textbf{Quality}. We applied commonly used FIDs \cite{chu2019neural} to evaluate the quality generated results, which are calculated through InceptionV3 by extracting the image features.

\textbf{Diversity}. We assess the method ability to create novel urban layouts for checking model collapse issuses.
For road networks, we calculate the Chamfer Distance (CD) against all paired road graphs \cite{chu2019neural}. 
For building layouts, we measure the degree of overlap between the generated images and the ground truth using the Mean Intersection over Union (mIoU) \cite{he2023globalmapper}. Higher CD in road networks generation and lower mIoU in building layouts generation indicates results with better diversity.

\textbf{Validity}. It is designed for geometry constraint generation tasks such as road network constraints and boundary constraints in building layouts generation tasks. Validity is the percentage of invalid samples, which are out-of-geometry constraints as shown in Fig. \ref{fig:metrics}.

\textbf{Urban Properties}. They are design to analysis the urban attributes of generated results. For road networks tasks, \textbf{Traffic Convenience} is the average value of $d_E (v_i, v_j) / d_S (v_i, v_j)$ for all node pairs $v_i, v_j$ that are over $300m$, where $d_E$ and $d_S$ are euclidean distance and Dijkstra shortest distance respectively, as shown in Fig. \ref{fig:metrics}. \textbf{Orientation} measures the entropy of street bearings \cite{boeing2019urban}.
Lower orientation indicates more ordered (such as grid-like) road networks.
For building layouts tasks, we calculate the \textbf{Wasserstein distance (WD)} for building height, counts and density to measure the distribution between the generated results and ground-truth datasets.

\subsection{Overall Results (RQ1,2)}
\begin{figure}[h]
  \centering
  \includegraphics[width=0.8\linewidth]{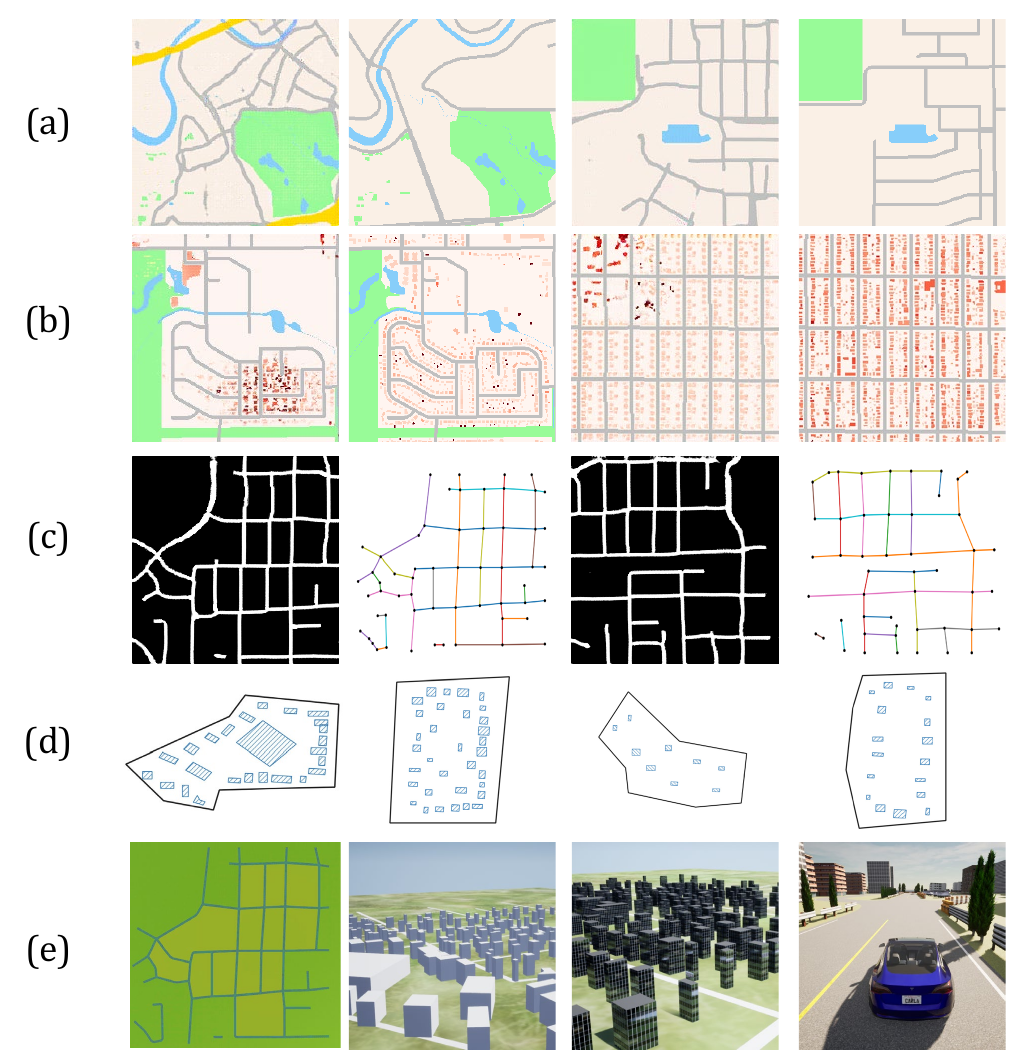}
  \caption{Examples of generated results. Row (a)$\sim$(d) are results from Task \uppercase\expandafter{\romannumeral1}$\sim$\uppercase\expandafter{\romannumeral4} respectively. The (e)th row is the visualization in UE of (d)th row.}
  \label{fig:results}
\end{figure}

We report the quantitative results for the four tasks in Tab. \ref{tab: results}, where "DIV" stands for diversity, "O" for road orientation, "C" for traffic convenience, "Ro" for road networks and "Bu" for building layouts generation tasks. "WD" measures the distribution of building counts, and "V" refers to validity. 
We divided the dataset into training, validation, and testing sets with ratios of 8:1:1, respectively.
The qualitative results are displayed in the (a)-th row of Figure \ref{fig:results}, where the first and third columns show the generated results, and the second and fourth columns present the ground truth. It is evident that the model used for Task \uppercase\expandafter{\romannumeral1} has successfully learned the pattern that roads should not cross mountains.
Compared Task \uppercase\expandafter{\romannumeral1} with \uppercase\expandafter{\romannumeral3}, both of which focus on road network generation, we conclude that Task \uppercase\expandafter{\romannumeral3} achieves superior image quality with a lower FID and higher road connectivity, but it also exhibits reduced diversity. 
Additionally, Task \uppercase\expandafter{\romannumeral3} is capable of generating more grid-like road networks when provided with corresponding attribute vectors.

For the building layout generation tasks (\uppercase\expandafter{\romannumeral2} and \uppercase\expandafter{\romannumeral4}), we conclude that the model proposed in Task \uppercase\expandafter{\romannumeral4} achieves much higher quality, evidenced by lower FID and WD. However, it tends to generate more buildings out of boundaries compared to the image-based model in Task \uppercase\expandafter{\romannumeral2}. The visualization of results for Tasks \uppercase\expandafter{\romannumeral2} and \uppercase\expandafter{\romannumeral4} is presented in the (b)-th and (d)-th rows of Fig. \ref{fig:results}, respectively. Additionally, the (d)-th row shows that building density attributes indeed controls the generative process.

\begin{table}
\centering
\caption{Quantitively results. Task \uppercase\expandafter{\romannumeral2} and \uppercase\expandafter{\romannumeral4} are building layouts generation while others are road networks generation.}
\label{tab: results}
\begin{tabular}{ccccc}
\toprule
Baseline    & FID$\downarrow$ & DIV(Ro/Bu)$\uparrow$           & Ro(O$\downarrow$/C$\uparrow$)   & Bu(WD/V)$\downarrow$ \\
\midrule
Task \uppercase\expandafter{\romannumeral1}               & 27.03        & 22.50/--                    & 3.27/0.75               & --          \\
Task \uppercase\expandafter{\romannumeral2}               & 32.73        &        --/\textbf{79.57}                  & --               & 7.64/\textbf{1.89}      \\
Task \uppercase\expandafter{\romannumeral4}             & \textbf{17.42}          & --/68.23 & --          & \textbf{3.37}/2.25    \\        
 \midrule
 Task\uppercase\expandafter{\romannumeral3}($\lambda_3$=0)          & 25.83       &   19.98/--                       & 3.25/0.77              & --      \\
Task\uppercase\expandafter{\romannumeral3}($\lambda_3$=0.3) & 26.98   & 19.05/--       & 3.27/\textbf{0.83}       & --  \\
RES\_5m       & 28.97   & 22.38/--       & 3.32/0.73       & --   \\
RES\_10m      & 35.28   & \textbf{24.24} /--      & 3.38/0.69       & --   \\
US\_1w      & 21.36   & 20.81/--       & \textbf{3.11}/0.80      & --   \\
Global\_1w  & 26.39   & 21.74/--       & 3.28/0.78      & --  \\
\bottomrule
\end{tabular}
\end{table}

\subsection{Ablation Studies (RQ2,4)}
We conduct ablation studies for road network generation to answer RQ2 and RQ4 in this subsection, as presented in Tab. \ref{tab: results}.
We conclude that the methods proposed in Task \uppercase\expandafter{\romannumeral3} greatly enhance connectivity by utilizing topological loss.
Additionally, we choose the 5m and 10m resolution (denoted "RES\_5m" and "RES\_10m" in Tab. \ref{tab: results}) as comparison, and conclude that as resolution changing from 5m to 10m, the quality of generated results worsens while diversity increases.
To explore the influence of dataset size, we randomly selected 10,000 samples from the entire dataset, denoted as "Global\_1w". We conclude that a smaller dataset size leads to higher diversity, compared to Task \uppercase\expandafter{\romannumeral1}. 
Additionally, we compared "Global\_1w" with "US\_1w", which indicates selected 10,000 samples within the United States. We conclude that a smaller collection area results in lower diversity, which underscores the importance of the diverse geographic coverage provided by the RoBus dataset.

\subsection{Applied to Auto Driving Simulations (RQ3)}
To validate the applicability of the RoBus dataset and proposed baselines in Task \uppercase\expandafter{\romannumeral3}  and \uppercase\expandafter{\romannumeral4}, we applied the generated results to autonomous simulation softwares such as \textit{CARLA} \cite{dosovitskiy2017carla}, which is built based on 3D game engine. 
As shown in last three rows of Fig. \ref{fig:results}, our pipeline for generating 3D urban scenes proceeds as follows. 
Initially, we generate road network graphics with a low orientation (grid-like road networks), using the methods proposed in Task \uppercase\expandafter{\romannumeral3}. For each city block boundary partitioned by road networks, we generate vectored building layouts with height using the methods outlined in Task \uppercase\expandafter{\romannumeral4}. 
We transfer the generated road networks to \textit{OpenDRIVE} format, starting by partitioning the road network graph into linestrings, which are then applied with CRS (WGS84). More importantly, we construct 3D buildings based on the generated building layouts with heights, and render the white building models in \textit{UnrealEngine} via randomly assigning different materials.

\section{Coclusion and Future Work}
In this work, we introduce the RoBus dataset, the first and largest open-source multimodal dataset designed for generative 3D urban design, specifically focusing on city-scale road networks and building layouts, which addresses the urgent need for comprehensive, high-quality training data for deep generative models.
To make it more applicable, we apply the generated 3D cities in UnrealEngine. 
However, the automated rendering of buildings is neglected in our work, which represents a promising topic for further research.
More importantly, there is substantial room for improvement on the generative model for generative 3D urban design, especially based on graphics like road networks topology and vectored buildings.
We expect the RoBus dataset and proposed baselines to inspire more creative and practical work in 3D city generation for multimedia games, metaverse and other socially aware multimedia applications.

\bibliographystyle{ACM-Reference-Format}
\bibliography{robus}

\appendix

\section{APPENDIX OVERVIEW}

\begin{itemize}
    \item Section \ref{sec:osmtags} introduces the key-value pairs used in OSM tags for extracting raw OSM data.
    \item Section \ref{sec:mgc} describes the algorithm developed to find the Geometric Minimal Cycle in a graph.
    \item Section \ref{sec:statics} presents additional statistical results from the RoBus dataset.
    \item Section \ref{sec:examples} provides further examples from the RoBus dataset.
    
\end{itemize}

\section{Data collection}
\label{sec:osmtags}
Roads plays a pivotal role in our dataset. We utilize road data sourced from OpenStreetMap (OSM) \footnote{\url{https://download.geofabrik.de/osm-data-in-gis-formats-free.pdf} \label{osmweb}} to acquire road segments and corresponding labels in OSM standards.
Specifically, we extract key-value pairs assigned to each "way". The key used to identify ways as roads is "\textbf{highway}", and the associated value specifies the type of roads. 
In our dataset, roads are divided into primary and secondary roads to support tasks like graded road networks generation (e.g., generating secondary roads based on the primary roads in a city). We conduct the categorization based on the values delineated in Tab. \ref{tab:osmtags}. For instance, to construct Road-P in our dataset, we extract roads in OSM which have the tag of {highway:motorway}, {highway:trunk}, or {highway:primary}. 

\begin{table}[h]
\centering
\caption{Extracting the following key-value pairs in OSM to our two road classes.}
\label{tab:osmtags}
\begin{tabular}{c|c|c}
                          & Road-P   & Road-S  \\ 
\hline
\multirow{4}{*}{highway} & motorway & secondary      \\
                          & trunck        & tertiary    \\
                          & primary        & residential   \\
                          &        & unclassified   \\
             
\end{tabular}
\end{table}

Similarly, we also extract water bodies and greenery spaces according to corresponding key-value pairs in OSM. 
To extract water bodies, we select the key 'water' with values 'reservoir' and 'river'. Additionally, we select the key 'natural' with values 'water', 'wetland', 'glacier' and the key 'leisure' with the value 'nature reserve'. We also select the key 'waterway' with values "riverbank", "dock", "canal", "drain", "ditch", "stream", "brook", "wadi", and "drystream".

To extract greenery spaces, we mainly select the key 'landuse' for values 'forest', 'farmland', 'allotments', 'meadow', "scrub", and "grass". To complete, we also select the key 'natural' with the value 'wood' and the key 'leisure' with the value 'garden'.

\section{Geometric Minimal Cycle}
\label{sec:mgc}
\begin{figure*}[h]
  \centering
  \includegraphics[width=\linewidth]{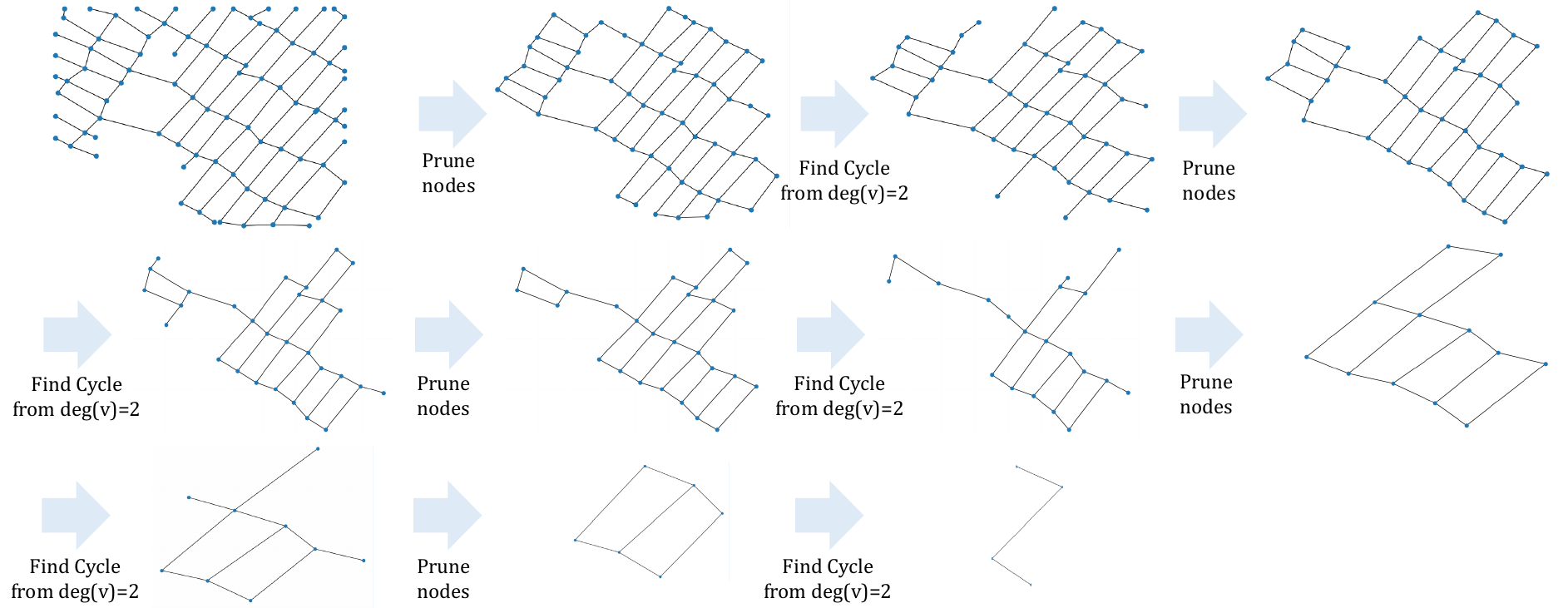}
  \caption{An Example of Finding Geometric Minimal Cycle. }
  \label{fig:gmc}
\end{figure*}

We employ automated partitioning of boundaries by road networks to generate building layouts at the block scale. A key challenge in this process is the efficient identification of \textit{geometric minimal cycles}, which are defined as the cycles in a graph with geometric coordinates that does not enclose any other cycles. 
The algorithm designed to find geometric minimal cycles is shown in Algorithm \ref{alg:gmc} and Fig. \ref{fig:gmc}.
Specifically,
We begin by removing all vertices with a degree of one from the graph.
Next, For each node with a degree of two, we identify simple paths connecting the node to its two immediate neighbors. 
Subsequently, we analyze all identified simple paths to determine the shortest cycles. 
We repeat the above steps iteratively until no more vertices remain in the graph.

\begin{algorithm}[!h]
\SetAlgoLined
\caption{Find Geometric Minimal Cycles in a Graph}
\label{alg:gmc}
\KwIn{Graph $G$ with vertices set $V$ and edges set $E$, the depth to stop search $cutoff$ }
\KwOut{The set $C$ of geometric minimal cycles }

\While{$G$ is not None}{
    \tcp{Remove all vertices with degree 1}
    \ForEach{vertex $v \in V$}{
        \If{deg($v$) == 1}{
            $G, V, E$ = RemoveVertex($v, G, V, E$)\;
        }
    }

    \tcp{Identify simple paths}
    
    \ForEach{$v \in V$ \textbf{and} deg($v$) == 2}{
        $v_1, v_2 \gets neighbour(v)$ \;
        \tcp{Find simple path from $v$ to $v_1$}
        $P \gets$ FindSimplePath($v, v_1, cutoff$)\;
        
        \tcp{Find geometric minimal cycles}
        $minc \gets None$ \;
        \ForEach{ $p \in P$ }{
            \If{$v_2 \in p$ }{
            $c \gets p + v$ \;
            \If{minc is None \textbf{or} len(c) $<$ len(minc)}{
                $minc \gets c$}
                }
            
        }
        \If{minc is not None}{$C$.append($minc$)\;
        $G, V, E$ = RemoveVertex($v, G, V, E$)\;
        }
    }
}

\end{algorithm}

\section{Dataset Statics}
\label{sec:statics}
\begin{figure}[h]
  \centering
  \includegraphics[width=0.7\linewidth]{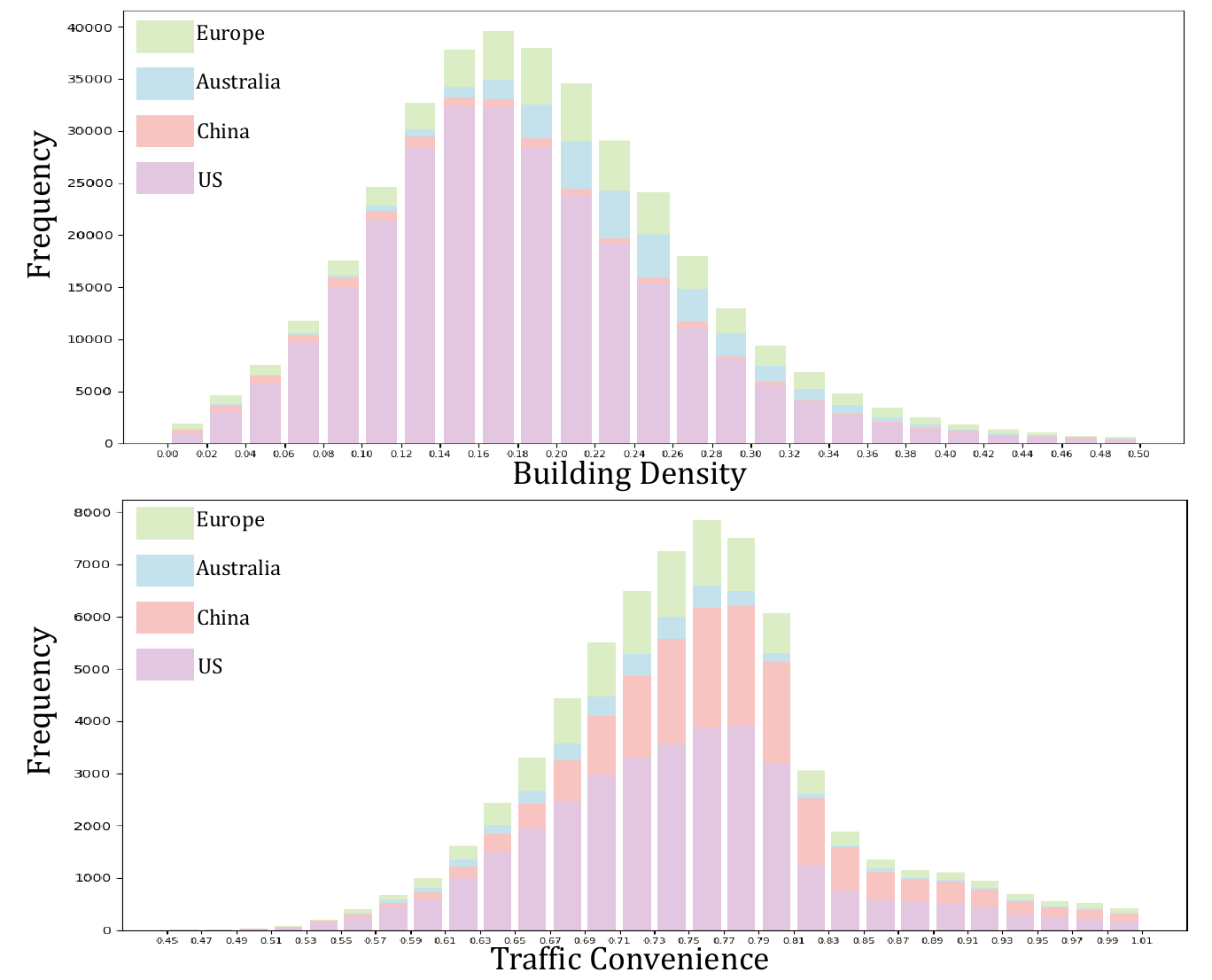}
  \caption{Statics of Building Density and Traffic Convenience. }
  \label{fig:statics2}
\end{figure}

We provide a more detailed statics of the RoBus dataset in Tab. \ref{tab:stats2} and Fig \ref{fig:statics2}.
As shown in Tab. \ref{tab:stats2}, the RoBus Dataset encompasses a total of 72,400 tiles, with the United States contributing the highest number at 37,429 tiles.
Australia's data in the dataset consists of 4,101 tiles, with roads totaling 39,771 km and an area coverage of 4,493 square kilometers.
China's data in the dataset includes 19,243 tiles, with road lengths of 123,227 km and an area coverage of 21,211 square kilometers.
The dataset includes 11,633 tiles from Europe, covering 12,960 square kilometers and including 105,450 km of roads.
For the United States, the dataset contains 37,429 tiles that cover 41,278 square kilometers and include 357,496 km of roads.
The RoBus Dataset's selection from different regions around the world ensures substantial \textbf{diversity}, reflecting a wide range of geographical variations.

\begin{table}[h]
\centering
\caption{Statistics of the RoBus Dataset by Country.}
\begin{tabular}{cccc}
  \toprule
   & \# of tiles      & Road Length ($km$)      & Covering ($km^2$)         \\ 
  \midrule
  Australia  & 4,101 & 39,771 & 4,493 \\
  China  & 19,243        & 123,227       & 21,211         \\
  Europe & 11,633          & 105,450      & 12,960        \\ 
  United States & 37,429          & 357,496       & 41,278        \\ 
  Total & 72,400          & 625,944       & 79,942        \\ 
  \bottomrule
\end{tabular}
\label{tab:stats2}
\end{table}

\section{RoBus Examples}
\label{sec:examples}
\begin{figure*}[h]
  \centering
  \includegraphics[width=\linewidth]{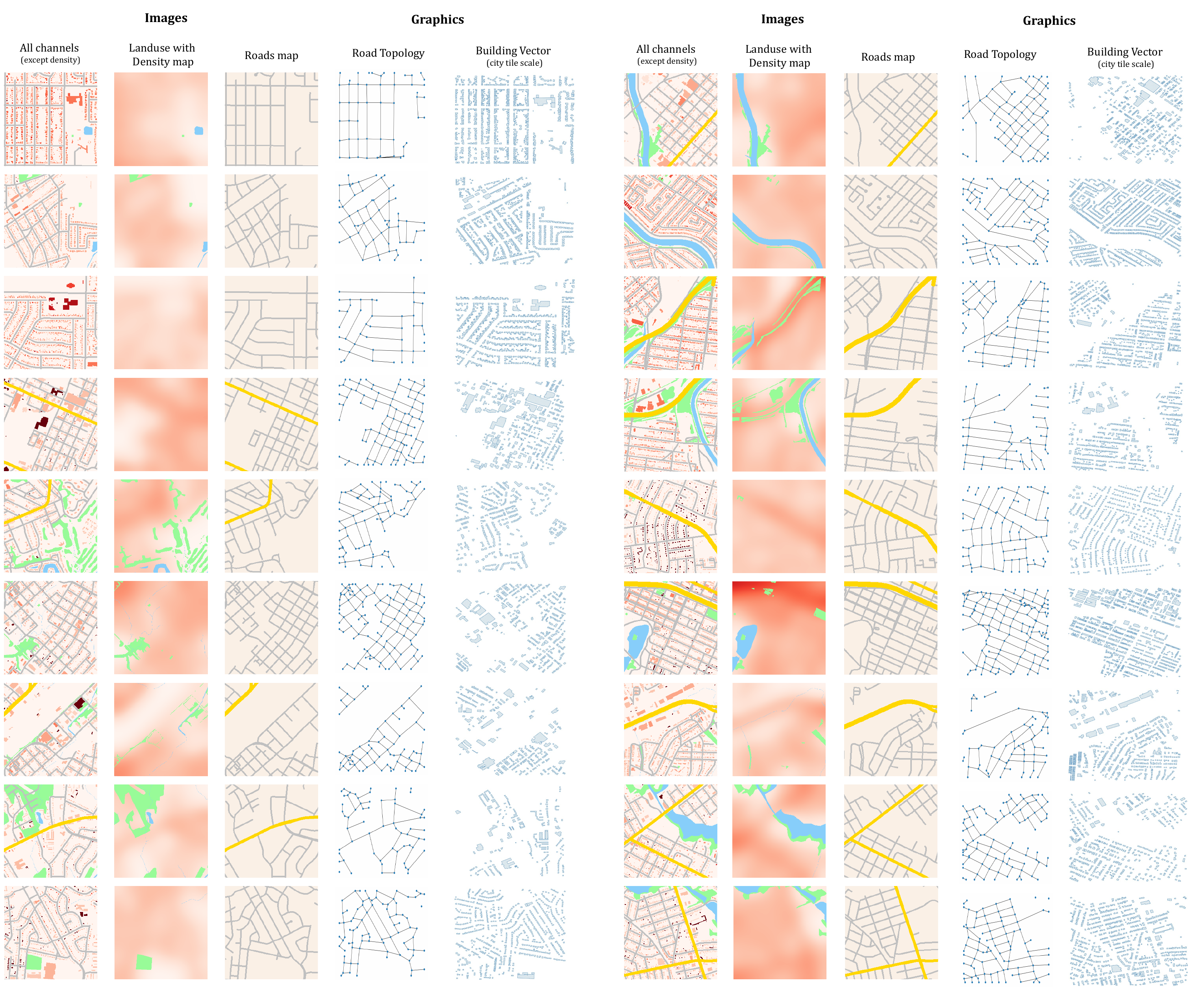}
  \caption{More Examples from RoBus Dataset. }
  \label{fig:examples}
\end{figure*}

We visualize the \textit{tif} images and graphics such as road topology (the fourth column) and building vectors (the fifth column) in Fig. \ref{fig:examples}. The visualization of tif images is structured into three columns, each highlighting different aspects of the data:
The first column displays images visualizing all channels except the density channel. The images are processed to exclude the density information, providing a clear view of the spatial distribution and characteristics of the other data layers.
The second column focuses on visualizing channels that represent water bodies, greenery spaces, and density. The images are crafted to specifically highlight these features, allowing for an immediate visual assessment of environmental and urban planning elements.
The visualization in the third column is dedicated to road channels. Here, primary roads (Road-P) are colored in golden, while secondary roads (Road-S) are colored in silver. This color coding not only distinguishes the two types of roads but also aids in understanding their hierarchy and connectivity within the road network.

Additionally, the original files are included in the supplementary directory '\textit{Samples\_from\_RoBus\_Dataset}'. For visualization and further analysis, users can utilize Python scripts or GIS software such as QGIS.

\end{document}